\documentclass{article}

\usepackage{PRIMEarxiv}

\usepackage[utf8]{inputenc} 
\usepackage[T1]{fontenc}    
\usepackage{hyperref}       
\usepackage{url}            
\usepackage{booktabs}       
\usepackage{amsfonts}       
\usepackage{nicefrac}       
\usepackage{microtype}      
\usepackage{lipsum}
\usepackage{fancyhdr}       
\usepackage{graphicx}       
\usepackage{amsmath}
\usepackage{multirow}

\DeclareMathOperator*{\argmin}{arg\,min}
\DeclareMathOperator*{\argmax}{arg\,max}

\graphicspath{{media/}}     

\usepackage{bm}
\usepackage{adjustbox}

\title{CORE: A Knowledge Graph Entity Type Prediction Method via Complex Space Regression and Embedding}

\author{
  Xiou Ge \\
  University of Southern California \\
  Los Angeles, USA\\
  \texttt{xiouge@usc.edu} \\
   \And
  Yun-Cheng Wang \\
  University of Southern California \\
  Los Angeles, USA\\
  \texttt{yunchenw@usc.edu} \\
   \And
  Bin Wang \\
  National University of Singapore \\
  Singapore\\
  \texttt{bwang28c@gmail.com} \\
   \And
  C.-C. Jay Kuo \\
  University of Southern California \\
  Los Angeles, USA\\
  \texttt{cckuo@sipi.usc.edu} \\
}

\begin{document}
\maketitle

\begin{abstract}
Entity type prediction is an important problem in knowledge graph (KG) research. A new KG entity type prediction method, named \textbf{CORE} (\textbf{CO}mplex space \textbf{R}egression and \textbf{E}mbedding), is proposed in this work. The proposed CORE method leverages the expressive power of two complex space embedding models; namely, RotatE and ComplEx models. It embeds entities and types in two different complex spaces using either RotatE or ComplEx. Then, we derive a complex regression model to link these two spaces. Finally, a mechanism to optimize embedding and regression parameters jointly is introduced. Experiments show that CORE outperforms benchmarking methods on representative KG entity type inference datasets. Strengths and weaknesses of various entity type prediction methods are analyzed. 
\end{abstract}


\section{Introduction}\label{sec:introduction}

Research on knowledge graph (KG) construction, completion, inference, and applications has grown rapidly in recent years since it offers a powerful tool for modeling human knowledge in graph forms. Nodes in KGs denote entities and links represent relations between entities. The basic building blocks of KG are entity-relation triples in form of (\emph{subject, predicate, object}) introduced by the Resource Description Framework (RDF). Learning representations for entities and relations in low dimensional vector spaces is one of the most active research topics in the field.

Entity type offers a valuable piece of information to KG learning tasks. Better results in KG-related tasks have been achieved with the help of entity type. For example, TKRL \cite{xie2016representation} uses a hierarchical type encoder for KG completion by incorporating entity type information. AutoETER \cite{Niu:AutoETER} adopts a similar approach but encodes the type information with projection matrices. Based on DistMult \cite{yang2014embedding} and ComplEx \cite{trouillon2016complex} embedding, \cite{jain2018type} propose an improved factorization model without explicit type supervision. JOIE \cite{hao2019universal} attempts to embed entities and types in two separate spaces by learning instance-view embedding and ontology-view embedding. Similar to JOIE, TaRP \cite{DBLP:conf/aaai/CuiKTGJ21} leverages the hierarchical type ontology structure for relation prediction. Instead of learning embedding for all types, TaRP develops a heuristic weighting mechanism to rank the prior probability of a relation given its head and tail type information. A similar idea was examined in TransT \cite{ma2017transt}. Besides, entity type is also important for Information Extraction tasks including entity linking \cite{gupta2017entity} and relation extraction \cite{zhou2005exploring,culotta2004dependency}.

\renewcommand{\figurename}{Figure}
\renewcommand{\tablename}{Table}

\begin{figure}[htbp]
\begin{minipage}[t]{0.5\linewidth}
    \includegraphics[width=\linewidth]{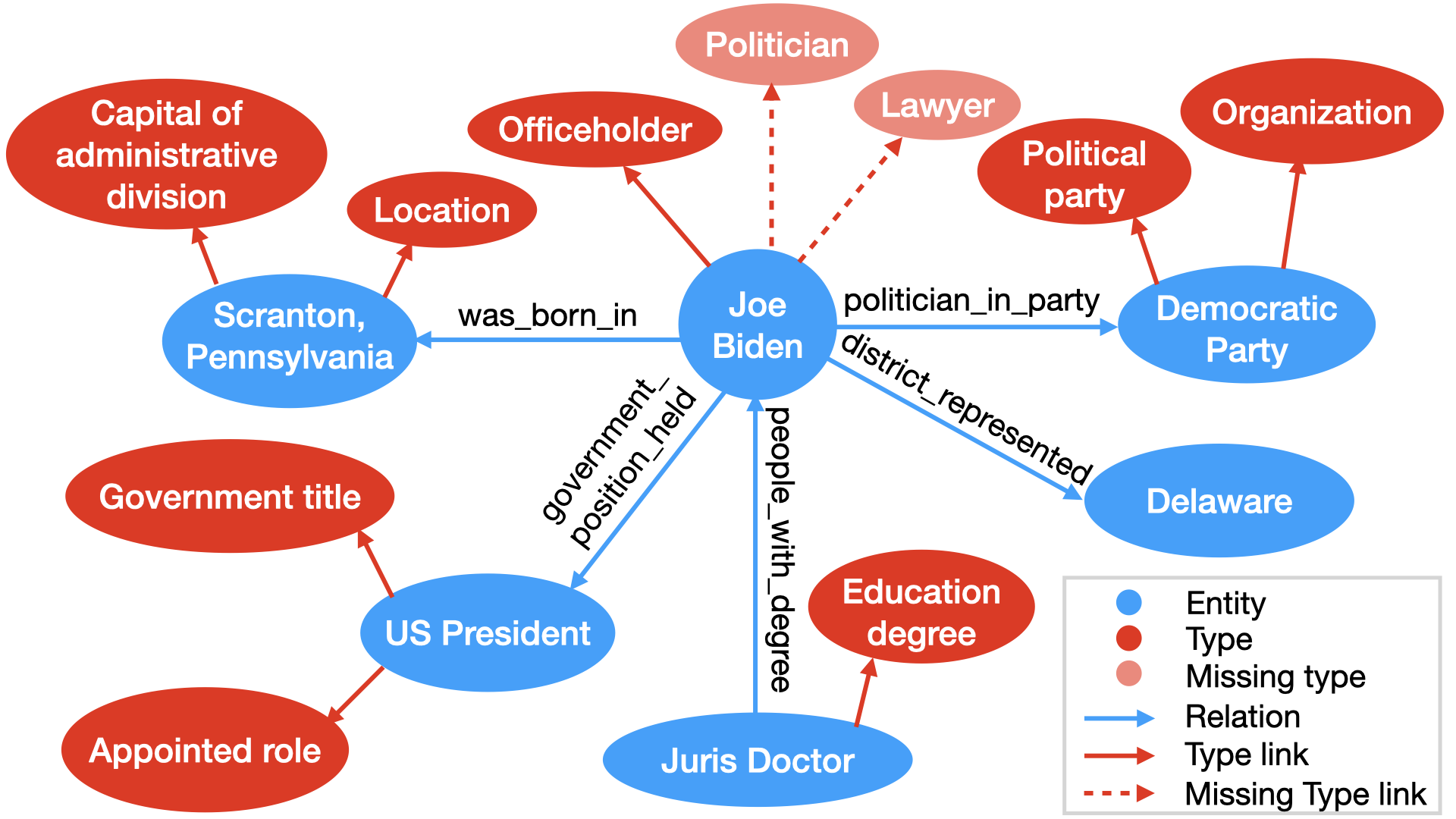}
    \caption{A KG with the entity type information.}
    \label{kge_illustration}
\end{minipage}%
    \hfill%
\begin{minipage}[t]{0.5\linewidth}
    \includegraphics[width=\linewidth]{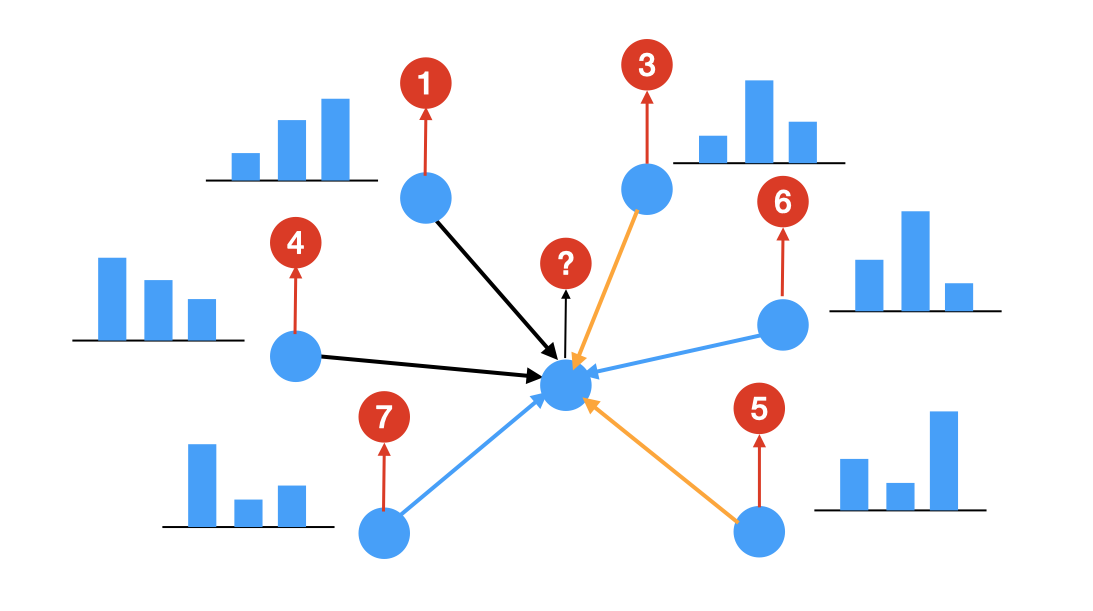}
    \caption{Illustration of the statistical approach.}
    \label{statistical_approach}
\end{minipage} 
\end{figure}

On the other hand, entity type prediction is challenging for several reasons. First, collecting additional type labels for entities is expensive. Second, type information is often incomplete especially for large-scale datasets. Fig. \ref{kge_illustration} shows a snapshot of a KG, where missing entity types need to be inferred. Third, KGs are ever-evolving and type information is often corrupted by noisy facts. Thus, there is a need to design algorithms to predict missing type labels. Quite a few approaches have been proposed to predict missing entity types in KG. They can be classified into three different categories: namely statistical-based, classifier-based, and embedding-based methods.  A brief review is given in Sec. \ref{sec:Related Works}. 

The contributions of our work are summarized as follows:
\begin{itemize}
    \item We present a new method for entity type prediction named CORE (COmplex space Regression and Embedding). CORE leverages the expressive power of complex space embedding models including RotatE \cite{sun2018rotate} and ComplEx \cite{trouillon2016complex} to represent entities and types. To capture the relatedness of entities and types, a complex regression model is built between entity space and type space.
    \item We conduct experiments on three major KG datasets and make performance comparison between CORE and state-of-the-art entity type prediction methods. CORE outperforms extant methods in most evaluation metrics.
    \item We study and compare statistical-based, classifier-based, and embedding-based methods for entity type prediction. Strengths and weaknesses of different approaches are discussed. We also introduce a better statistical method baseline named SDType-Cond.
\end{itemize}

\section{Related Work}\label{sec:Related Works}
Work on entity type prediction can be categorized into three types as elaborated below.

\subsection{Statistical Approach.} Before machine learning is introduced to KG entity type prediction, the type inference is often performed using RDF rules and graph pattern matching \cite{gangemi2012automatic}. Although these handcrafted rules can make type prediction with high precision, they tend to miss a lot of possible cases since it is almost impossible to manually create all patterns. Hence, this approach is not scalable for large scale datasets. As such, researchers apply basic statistics to solve the KG entity type prediction problem. An early method, called SDType \cite{paulheim2013type}, predicts missing entity types by estimating the empirical probability distribution of $P(type|relation)$ and aggregating all such conditional probabilities generated by neighboring relations of the target entity. Although this approach is robust to noise, it cannot predict unseen types and relation combinations. Another shortcoming of the statistical approach is that its performance deteriorates as the number of entity types becomes larger (say, with thousands of entity types). Furthermore, it does not exploit the type information of neighboring entities. In this paper, we will show that the type prediction performance can be further improved by conditioning on the type information of neighboring entities. Fig. \ref{statistical_approach} illustrates the statistical approach for entity type prediction. Suppose we want to predict the type of the target node, then given the information of each neighbor relation and neighbor node's type, we can estimate the type distribution of for the target node.

\subsection{Classification Approach.} Node classification is a common task in graph learning. One solution is to train a classifier with node features as the input and corresponding labels as the output.  This idea can be applied to entity type prediction as well. Researchers \cite{yaghoobzadeh-schutze-2015-corpus, xin2018improving, sofronova2020entity} conducted experiments with entity textual descriptions in form of word embeddings as features and neural networks such as MLP, LSTM, and CNN as classifiers.  An end-to-end architecture that learns entity embedding and type prediction jointly was proposed in \cite{jin2018attributed}. The correlation between entity attributes and KG link structures were taken into account in the learning of distributed entity representations. In this work, with pretrained KG embeddings as features, we test a couple of classifiers such as SVM and XGBoost \cite{chen2016xgboost}. Again, we find this approach does not perform well for large datasets. In fact, a comparative study on statistical and classifier approaches was conducted in \cite{jain2021embeddings}.  By analyzing results from selected entity type classes, they concluded that KG embeddings fail to capture the semantics of entities and statistical approaches are often superior in type prediction. While we concur with their experimental findings that the combination of entity embedding features and classifiers tend to yield poor results, we do not agree that KG embedding models cannot be useful for entity type prediction as elaborated below. 

\subsection{Embedding Approach.} Before introducing the embedding approach for entity type prediction, it is worthwhile to review KG Embedding models briefly. According to \cite{ji2021survey}, a KG embedding model can be categorized based on its representation space and scoring function. Among several representation spaces, real and complex vector spaces are the two most common ones. TransE models triples in form of \emph{subject, relation, object} or $(s,r,o)$ in $d$-dimensional real vector space with the translational principle $\mathbf{e_s}+\mathbf{w_r}\approx\mathbf{e_o}$. Yet, TransE is not suitable for modeling asymmetric and many-to-one relations. To overcome this weakness, researchers venture into the complex vector space for and design models with greater expressive power.  ComplEx and RotatE are two prominent examples. ComplEx is motivated by low rank matrix factorization in the complex space to model both symmetric and asymmetric relations effectively. Inspired by Euler's identity, RotatE models relations as rotations in the complex vector space to remedy ComplEx's inability to model composition patterns in KG. Both TransE and RotatE adopt distance-based scoring function while ComplEx has a semantic matching score. 

Besides KG embedding, embedding-based entity type prediction approaches
learn a distributed representation of entity types. For example, a
distance-based scoring function can be used to measure the relevance of
a particular type to the target entity.  The ETE model
\cite{moon2017learning} embeds entity and entity types in the same
space. ConnectE \cite{zhao-etal-2020-connecting} embeds
entities and types in two different spaces and learns a mapping from the
entity space to the type space. It leverages neighbor label information
to boost the performance further.  Based on a similar idea, JOIE
\cite{zhao-etal-2020-connecting} adds an intra-view component to model
the hierarchical structure of the type ontology. ETE, ConnectE, and JOIE all adopt TransE embedding to represent KG entities.  JOIE targets at better
results for link prediction whereas ETE and ConnectE focus on improving
entity type prediction. TransE is known to suffer from a few problems, e.g.,
not able to model asymmetric relations. Poor relation representation
leads to poor entity representation.  Since the quality of entity
representation affects entity type prediction performance, we exploit
the expressive power of complex-space KG embedding to achieve better
results. 

\section{Proposed CORE Methods}\label{sec:method}

To leverage the expressive power of the complex-space KG embedding, we
propose to learn a set of embeddings for both the entity space and the
type space as shown in Fig. \ref{space_illustration}.  Specifically, we experiment with ComplEx and RotatE embedding models. On top of entity embedding and type embedding, we learn a regression between these two spaces. Finally, we make type predictions using a distance-based scoring scoring function based on the embeddings and regression parameters. The high-level concept of the proposed CORE model is given in Fig. \ref{CORE_model}.

\subsection{Complex Space KG Embedding.} Let $(s, r, o)$ be a KG
triple and $\mathbf{e_s},\mathbf{w_r},\mathbf{e_o}\in\mathbb{C}^k$
denote the complex space representation of triple's subject, relation,
and object. For ComplEx, the score function is
\begin{align*}
f_r(s,o)=-\operatorname{Re}(\langle \mathbf{w_r}, \mathbf{e_s}, 
\mathbf{\overline{e}_o} \rangle),
\end{align*}
where $\langle\cdot,\cdot,\cdot\rangle$ denotes an element-wise
multi-linear dot product, and $\overline{\ \cdot\ }$ denotes the
conjugate for complex vectors.  For RotatE embedding, the score function is
\begin{align*}
f_r(s,o)=\|\mathbf{e_s}\circ \mathbf{w_r}-\mathbf{e_o}\|_1,
\end{align*}
where $\circ$ denotes the element-wise product.

We follow RotatE's negative sampling loss and self-adversarial training strategy to train the embedding. The loss function for KG embedding models can be written as
\begin{align*}
     L_{\textnormal{KGE}}=-\log\sigma(\gamma_1-f_r(s, o))-\sum_{i=1}^np(s'_i,r,o'_i)\log\sigma(f_r(s'_i,o'_i)-\gamma_1),
\end{align*}
where $\sigma$ is the sigmoid function, $\gamma_1$ is a fixed margin
hyperparameter for training KG embedding, $(s'_i,r,o'_i)$ is the $i$th
negative triple and $p(s'_i,r,o'_i)$ is the probability of drawing
negative triple $(s'_i,r,o'_i)$. Given a positive triple, $(s_i,r,o_i)$,
the negative sampling distribution is
\begin{align*}
p(s'_j,r,o'_j|\{(s_i,r,o_i)\})=\frac{\exp \alpha_1 f_r(s'_j, o'_j)}
{\sum_i \exp \alpha_1 f_r(s'_i, o'_i)},
\end{align*}
where $\alpha_1$ is the temperature of sampling.

\subsection{Complex Space Type Embedding.} Similar to definitions in KG
Embedding space, we use $(s_t, r, o_t)$ to
denote a type triple and $\mathbf{t_s},\mathbf{v_r},\mathbf{t_o}
\in\mathbb{C}^l$ to denote representations of the subject type, the
relation, and the object type in the type embedding space.
For ComplEx embedding, the score function is
\begin{align*}
f_r(s_t,o_t)=-\operatorname{Re}(\langle
\mathbf{v_r}, \mathbf{t_s}, \mathbf{\overline{t}_o} \rangle).
\end{align*}

For RotatE embedding, the score function is
\begin{align*}
f_r(s_t,o_t)=\|\mathbf{t_s}\circ 
\mathbf{v_r}-\mathbf{t_o}\|_1.
\end{align*}

To train type embedding, we use the self-adversarial negative sampling 
loss in form of
\begin{align*}
L_{\textnormal{TPE}}=-\log\sigma(\gamma_2-f_r(s_t,o_t))-\sum_{i=1}^np(s_{t_i}',r,o_{t_i}')\log 
\sigma(f_r(s_{t_i}',o_{t_i}')-\gamma_2),
 \end{align*}
where $\gamma_2$ is a fixed margin hyperparameter for training type
embedding and $(s_{t_i}',r,o_{t_i}')$ is the
$i$th negative triple, and $p(s_{t_i}',r,o_{t_i}')$ is
the probability of drawing the negative triple
$(s_{t_i}',r,o_{t_i}')$. Given a
positive triple, $(s_{t_i},r,o_{t_i})$, the
negative sampling distribution is
\begin{align*}
p(s_{t_j}',r,o_{t_j}'|\{(s_{t_i},
r,o_{t_i})\})=\frac{\exp \alpha_2f_r(s_{t_j}',
o_{t_j}')}{\sum_i \exp \alpha_2 f_r(s_{t_i}',
o_{t_i}')},
\end{align*}
where $\alpha_2$ is the temperature of sampling.

\begin{figure}[htbp]
\begin{minipage}[t]{0.5\linewidth}
    \includegraphics[width=\linewidth]{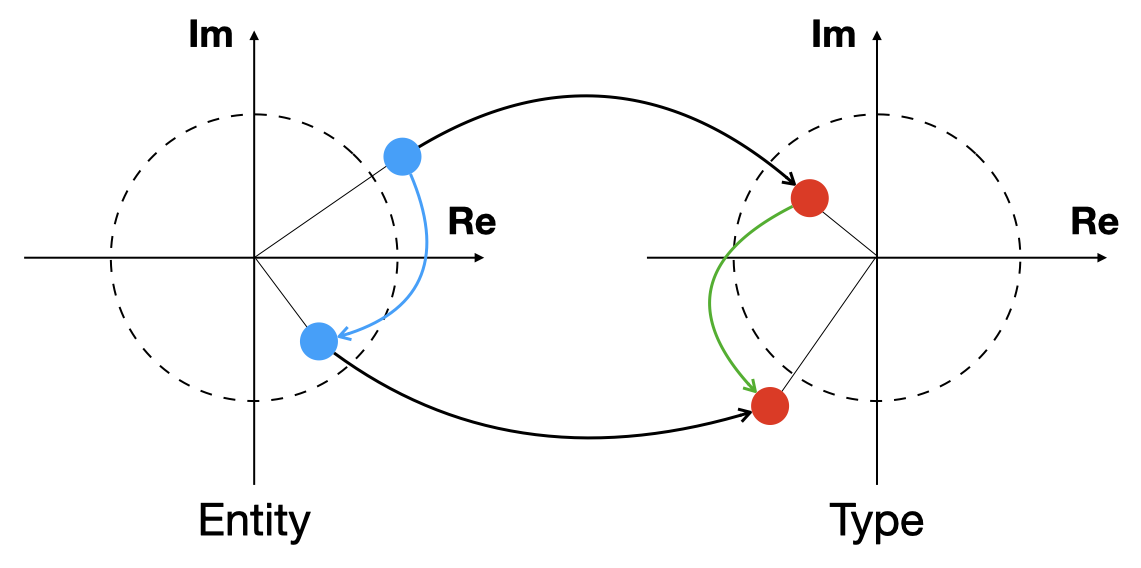}
    \caption{Illustration of RotatE entity space and RotatE type space and the regression linking two spaces.}
    \label{space_illustration}
\end{minipage}%
    \hfill%
\begin{minipage}[t]{0.45\linewidth}
    \includegraphics[width=\linewidth]{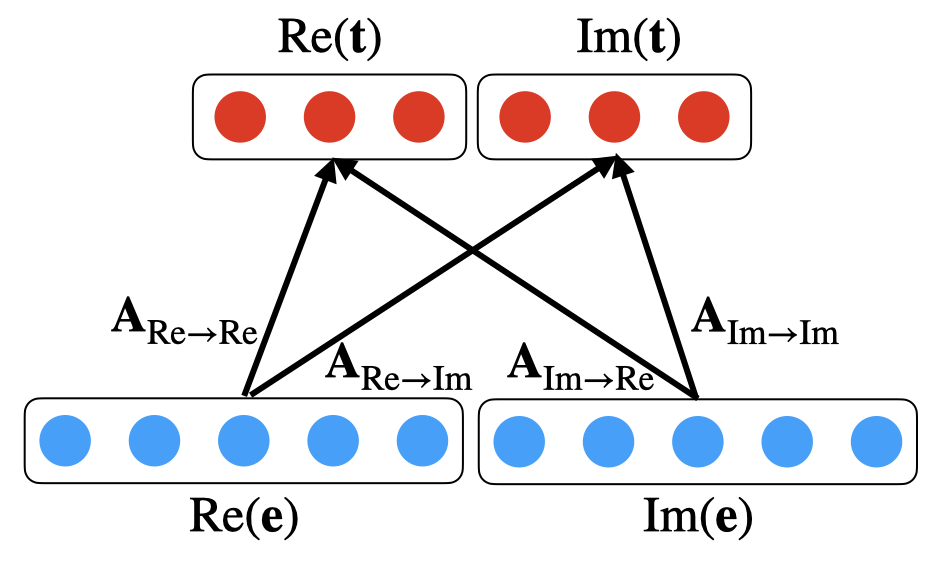}
    \caption{Illustration of the CORE Model, where the blue and red dots
denote entities and types in their complex embedding spaces,
respectively.}
    \label{CORE_model}
\end{minipage} 
\end{figure}

\subsection{Solving Complex Space Regression.} To propagate
information from the entity space to the type space, we learn a regression between two complex spaces. A feasible and logical way of solving the complex regression is to cast the problem into a multivariate regression problem in real vector space. Formally, let $\mathbf{e}\in\mathbb{C}^k$, $\mathbf{t}\in\mathbb{C}^l$ denote the representation of the entity and its type. We divide the real and the imaginary parts of every complex entity vector into two real vectors; namely, $\operatorname{Re(\mathbf{e})}\in \mathbb{R}^k$ and $\operatorname{Im(\mathbf{e})}\in \mathbb{R}^k$. We do the same to divide the complex type vector into two real vectors: $\operatorname{Re(\mathbf{t})}\in \mathbb{R}^l$ and $\operatorname{Im(\mathbf{t})}\in \mathbb{R}^l$. 

As shown in Fig. \ref{CORE_model}, the regression process consists of four
different real block matrices: $\{\mathbf{A}_{\operatorname{Re}
\rightarrow \operatorname{Re}}$, $\mathbf{A}_{\operatorname{Im}
\rightarrow \operatorname{Re}}$, $\mathbf{A}_{\operatorname{Re}
\rightarrow \operatorname{Im}}$ and $\mathbf{A}_{\operatorname{Im}
\rightarrow \operatorname{Im}}\}\in\mathbb{R}^{k\times l}$. 
The real part of the output vector depends on both the real
and imaginary part of the input vector. Similarly, the imaginary part
of the output vector also depends on both the real and imaginary
part of the input vector. The regression problem can
be rewritten as
\begin{align*}
    \begin{pmatrix}
        \operatorname{Re(\mathbf{t})} \\\
        \operatorname{Im(\mathbf{t})}
    \end{pmatrix} =
    \begin{pmatrix}
    \mathbf{A}_{\operatorname{Re} \rightarrow \operatorname{Re}} & \mathbf{A}_{\operatorname{Im} \rightarrow \operatorname{Re}}\\\
    \mathbf{A}_{\operatorname{Re} \rightarrow \operatorname{Im}} & \mathbf{A}_{\operatorname{Im} \rightarrow \operatorname{Im}}
    \end{pmatrix} \begin{pmatrix}
        \operatorname{Re(\mathbf{e})} \\\
        \operatorname{Im(\mathbf{e})}
    \end{pmatrix} + \mathbf{\epsilon},
\end{align*}
where $\mathbf{\epsilon}$ denotes the error vector. To minimize
$\mathbf{\epsilon}$, we use the following score function
\begin{align*}
f(e,t) &= \|\mathbf{A}_{\operatorname{Re} \rightarrow \operatorname{Re}}\cdot 
\operatorname{Re(\mathbf{e})} + \mathbf{A}_{\operatorname{Im} \rightarrow 
\operatorname{Re}} \cdot \operatorname{Im(\mathbf{e})} - \operatorname{Re(\mathbf{t})}\|_2 \\ 
&+\|\mathbf{A}_{\operatorname{Re} \rightarrow \operatorname{Im}}\cdot 
\operatorname{Re(\mathbf{e})} + \mathbf{A}_{\operatorname{Im} \rightarrow 
\operatorname{Im}} \cdot \operatorname{Im(\mathbf{e})} - \operatorname{Im(\mathbf{t})}\|_2.
\end{align*}

We find that the self-adversarial negative sampling strategies are
useful in optimizing regression coefficients. The loss function in
learning these coefficients is set to
\begin{align*}
L_{\textnormal{RR}}=-\log\sigma(\gamma_3-f(e,t))-\sum_{i=1}^np(e,t'_i)
\log\sigma(f(e,t'_i)-\gamma_3),
\end{align*}
where $\gamma_3$ is a fixed margin hyperparameter in the regression,
$(e,t'_i)$ is the $i$th negative pair, and $p(e,t'_i)$ is the
probability of drawing negative pair $(e,t'_i)$. Given positive triple
$(e,t_i)$, the negative sampling distribution is equal to
\begin{align*}
p(e,t'_j|\{(e,t_i)\})=\frac{\exp \alpha_3 f(e, t'_j)}{\sum_i 
\exp \alpha_3 f(e, t'_i)},
\end{align*}
where $\alpha_3$ is the temperature of sampling.

\subsection{Type Prediction.} 
We use the distance-based scoring function $f(e,t)$ in the regression to predict entity types. The type prediction function can be written as
\begin{align*}
\hat{t} = \argmin_{t \in \mathcal{T}} f(e,t),
\end{align*}
where $\mathcal{T}$ denotes the set of all types.

\subsection{Optimization.} We first initialize the embedding and
regression parameters by sampling from the standard uniform
distribution. Three parts of our model are optimized sequentially.
First, we optimize the KG embeddings using KG triples and negative
triples. Next, we move on to train regression and type space embeddings
parameters. we freeze the KG embedding to ensure the regression is
learning important information in this stage.  Last, we further optimize
the type space embeddings using type triples. To avoid overfitting of
the regression model in the early training stage, we alternate the
optimization for each part of the model every 1000 iterations.

\subsection{Complexity.} The memory and space complexity for CORE are both $O(n_e \times d_e + n_r \times d_r + n_t \times d_t + d_e \times d_t)$, where $n$ denotes the number of objects, $d$ denotes the dimension, and the subscripts $e$ denotes entity, $r$ denotes relation, $t$ denotes type, respectively.

\begin{table*}[htb]
\centering
\begin{tabular}[t]{c|c|c|c|c|c|c|c|c|c}
\hline
\multirow{2}{*}{\bf Dataset} & \multirow{2}{*}{\bf \#Ent} & \multirow{2}{*}{\bf \#Rel } & \multirow{2}{*}{\bf \#Type } & \multicolumn{3}{c|}{\bf \#KG Triples} & \multicolumn{3}{c}{\bf \#Entity Type Pairs}\\\cline{5-10}
& & & &{\bf\#Train }& {\bf \#Valid} &{\bf \#Test } &{\bf\#Train }& {\bf \#Valid} &{\bf \#Test }\\ \hline
 {FB15k-ET}  & 14,951 & 1,345 & 3,851 & 483,142 & 50,000 & 59,071 & 136,618 & 15,749 & 15,780 \\
 {YAGO43k-ET}  & 42,335 & 37 & 45,182 &331,687 & 29,599 & 29,593 & 375,853 & 42,739 & 42,750 \\
 {DB111K-174}  & 111,762 & 305 & 242 & 527,654 & 65,000 & 65,851 & 57,969 & 1,000 & 39,371  \\ \hline
\end{tabular}
\caption{Statistics of three KG datasets used in our experiments.}\label{data_stats}
\end{table*}

\section{Experiments}\label{sec:experiments}

\subsection{Datasets}

We evaluate the proposed CORE model by conducting experiments on several
well-known KG datasets with the entity type information.  They include
FB15k-ET, YAGO43k-ET \cite{moon2017learning}, and DB111k-174
\cite{hao2019universal}, which are subsets of Freebase
\cite{bollacker2008freebase}, YAGO \cite{suchanek2007yago}, and
DBpedia \cite{auer2007dbpedia} KGs, respectively.
\cite{zhao-etal-2020-connecting} further clean FB15k-ET and YAGO43k-ET
datasets by removing triples in the training sets from the
validation and test sets. They also create a script to generate type triples $(s_t, r, o_t)$ by enumerating all possible combination of relations and the types of their subject and object. We use the same script to generate type triples for training type embedding. The statistics of these datasets are shown in Table
\ref{data_stats}. 

\subsection{Hyperparameter Setting} 

\begin{figure}[htbp]
\begin{minipage}[t]{0.45\linewidth}
    \includegraphics[width=\linewidth]{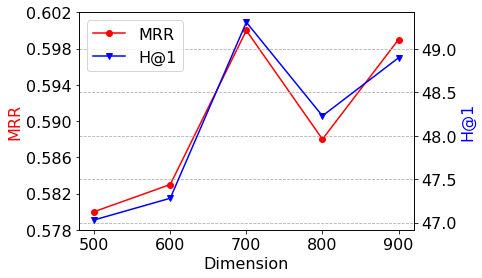}
    \caption{Comparison of the MRR performance for FB15k-ET as a function of 
the type dimension with RotatE embedding.}
    \label{RotatE_dim}
\end{minipage}%
    \hfill%
\begin{minipage}[t]{0.45\linewidth}
    \includegraphics[width=\linewidth]{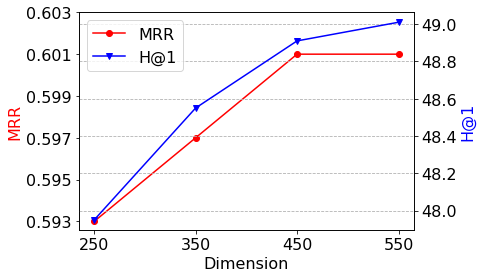}
    \caption{Comparison of the MRR performance for FB15k-ET as a function of 
the type dimension with ComplEx embedding.}
    \label{ComplEx_dim}
\end{minipage} 
\end{figure}

We list out the hyperparameter settings for each of the benchmarking datasets we run experiments on in Table ~\ref{hyperparameter_setting}. In this table, $k$ and $l$ denote the dimension of entity embedding and type embedding, respectively. $Ebz$, $Tbz$, and $Nsz$ denote the entity batch size, type batch size, and negative sample size, respectively. $\alpha_1$, $\gamma_1$, and $\eta_1$ denote the sampling temperature, margin parameter, and learning rate respectively. In addition, we also show the MRR and Hits@k results for RotatE and ComplEx with different type dimensions for FB15k-ET in Fig.  \ref{RotatE_dim} and Fig.
\ref{ComplEx_dim}, respectively.

\begin{table*}[htb]
\centering
\begin{tabular}[t]{c|c|c|c|c|c|c|c|c|c}
\hline
 \bf Dataset                & \bf Model & $k$ & $l$ & $Ebz$ & $Tbz$ & $Nsz$ & $\alpha_1$ & $\gamma_1$ & $\eta_1$ \\\hline
 \multirow{2}{*}{FB15k-ET}  & CORE-RotatE   & 1000 & 700 & 1024  & 4096  & 256   & 1        & 24       & 0.0001\\\cline{2-10}
                            & CORE-ComplEx  & 500  & 550 & 1024  & 4096  & 400   & 1        & 24       & 0.0002\\\hline
 \multirow{2}{*}{YAGO43k-ET}  & CORE-RotatE   & 500 & 350 & 1024  & 4096  & 400   & 1        & 24       & 0.0002\\\cline{2-10}
                            & CORE-ComplEx  & 500  & 350 & 1024  & 4096  & 400   & 1        & 24       & 0.0002\\\hline
 \multirow{2}{*}{DB111K-174}  & CORE-RotatE   & 1000 & 250 & 1024  & 4096  & 256   & 1        & 24       & 0.0005\\\cline{2-10}
                            & CORE-ComplEx  & 1000  & 250 & 1024  & 4096  & 400   & 1        & 24       & 0.0002\\\hline
\end{tabular}
\caption{Hyperparameter setting.}\label{hyperparameter_setting}
\end{table*}

\begin{table*}[ht]
  \begin{center}
    \begin{tabular}{c|cccc|cccc} 
      \hline
      \textbf{Datasets} && \multicolumn{2}{c}{\textbf{FB15k-ET}} &&& \multicolumn{2}{c}{\textbf{YAGO43k-ET}} &\\
      \hline
      \textbf{Metrics} & \textbf{MRR} & \textbf{H@1} & \textbf{H@3} & \textbf{H@10}& \textbf{MRR} & \textbf{H@1} & \textbf{H@3} & \textbf{H@10}\\
      \hline
      RESCAL \cite{nickel2011three} & 0.19&9.71&19.58&37.58 &0.08&4.24&8.31&15.31\\
      RES.-ET \cite{moon2017learning}  &0.24&12.17&27.92&50.72 &0.09&4.32&9.62&19.40\\
      HOLE \cite{nickel2016holographic} &0.22&13.29&23.35&38.16&0.16&9.02&17.28&29.25\\
      HOLE-ET \cite{moon2017learning} &0.42&29.40&48.04&66.73 &0.18&10.28&20.13&34.90\\
      TransE \cite{bordes2013translating}  &0.45&31.51&51.45&73.93&0.21&12.63&23.24&38.93\\
      TransE-ET \cite{moon2017learning} &0.46&33.56&52.96&71.16&0.18&9.19&19.41&35.58\\
      ETE \cite{moon2017learning} &0.50&38.51&55.33&71.93&0.23&13.73&26.28&42.18\\
      ConnectE-E2T \cite{zhao-etal-2020-connecting}  & 0.57 & 45.53 & 62.31 & 78.12 & 0.24 & 13.54 & 26.20 & 44.51\\
      ConnectE-E2T-TRT \cite{zhao-etal-2020-connecting} & 0.59 & \textbf{49.55} & 64.32 & 79.92 & 0.28 & 16.01 & 30.85 & 47.92\\
      ConnectE-E2T-TRT (Actual) & 0.58 & 47.45 & 64.33 & 77.55 & 0.14 & 8.04 & 17.59 & 24.62\\
      \hline
      SDType-Cond & 0.42 & 27.56 & 50.09 & 71.23 & - & - & - & -\\
      CORE-RotatE & \textbf{0.60} & \underline{49.32} & \underline{65.25} & \underline{81.09} & \underline{0.32} & \underline{22.96} & \underline{36.55} & \underline{51.00}\\
      CORE-ComplEx & \textbf{0.60} & 48.91 & \textbf{66.30} & \textbf{81.60} & \textbf{0.35} & \textbf{24.17} & \textbf{39.18} & \textbf{54.95}\\
      \hline
    \end{tabular}
\caption{Performance comparison of various entity type prediction
methods in terms of filtered ranking for FB15k-ET and
YAGO43k-ET, where the best and the second best performance 
numbers are shown in bold face and with an underscore, respectively.}\label{fbyago}
\end{center}
\end{table*}

\begin{table}[ht]
  \begin{center}
    \begin{tabular}{c|ccc} 
      \hline
      \textbf{Datasets} && \textbf{DB111K-174} \\
      \hline
      \textbf{Metrics} & \textbf{MRR} & \textbf{H@1} & \textbf{H@3}\\
      \hline
      JOIE-HATransE-CT & 0.857 & 75.55 & 95.91\\
      
      SDType & 0.861 & 78.53 & 92.67 \\
      ConnectE-E2T & 0.88 & 81.63 & 94.19\\
      ConnectE-E2T-TRT & \underline{0.90} & 82.96 & \underline{96.07}\\
      \hline
      SDType-Cond & 0.879 & 80.99 & 94.05\\
      CORE-RotatE & 0.889 & 82.02 & 95.36 \\
      CORE-ComplEx & \underline{0.900} & \underline{84.25} & 95.42 \\
      TransE+XGBoost &  0.878 & 81.38 & 94.07\\
      TransE+SVM & \textbf{0.917} & \textbf{86.77} & \textbf{96.33}\\
      \hline
    \end{tabular}
\caption{Performance comparison of entity type prediction for
DB111K-174, where the best and the second best performance numbers are
shown in bold face and with an underscore, respectively.}\label{DB111K174}
\end{center}
\end{table}

\begin{table*}[ht]
  \begin{center}
    \begin{tabular}{c|c|c} 
      \hline
      \textbf{Entity} & \textbf{Model} & \textbf{Top 3 Type Predictions} \\
      \hline
      \multirow{3}{*}{Albert Einstein} & ConnectE & Islands of Sicily, Swiss singers, Heads of state of Canada \\\cline{2-3}
       & \multirow{2}{*}{CORE} & \textbf{Nobel laureates in Physics}, Fellows of the Royal Society, \\
       & & 20th-century mathematicians\\
      \hline
       \hline
       \multirow{3}{*}{Warsaw} & \multirow{2}{*}{ConnectE} & Defunct political parties in Poland, Political parties in Poland,\\
       & & Universities in Poland \\\cline{2-3}
     & CORE & \textbf{Administrative district}, Port cities, Cities in Europe\\
      \hline
      \hline
      \multirow{3}{*}{George Michael} & \multirow{2}{*}{ConnectE} & United Soccer League players, People from Stourbridge,\\
      & & Fortuna Düsseldorf managers \\\cline{2-3}
      & CORE & \textbf{British singers}, English musicians, Rock singers\\
      \hline
    \end{tabular}
\caption{An illustrative example of type prediction for the YAGO43k-ET 
dataset.}\label{actual_type_prediction}
\end{center}
\end{table*}

\subsection{Benchmarking Methods} 

To the best of our knowledge, we are the first work to compare embedding-based methods with statistical-based and classifier-based methods since we would like to understand the strengths and weaknesses of different models. 

\subsubsection{Statistical-based Method.} We compare the performance of
the proposed CORE model with that of a statistical-based method named SDType-Cond in
FB15k-ET dataset.  SDType-Cond is a variant of SDType. The neighbor
type is readily available in many entities, yet SDType ignores this
important piece of information.  SDType-Cond is capable of estimating
the type distribution more precisely by leveraging the known neighbor
type information. Instead of estimating the type distribution given a
paricular relation $p(s_t|r=\Tilde{r})$ or
$p(o_t|r = \Tilde{r})$, we estimate $p(s_t|r =
\Tilde{r}, o_t = \Tilde{t})$ or $p(o_t|r =
\Tilde{r}, s_t = \Tilde{t})$, $\Tilde{r}\in \mathcal{R}$,
$\Tilde{t}\in\mathcal{T}$, where $\mathcal{R}$ and $\mathcal{T}$ denote
the set of all relations and the set of all entity types, respectively.
The two probabilities represent the two cases where the target entity
serves as a subject and an object, respectively. By aggregating the
probabilities generated by all possible $(\Tilde{r}, \Tilde{t})$
combinations in the neighborhood of the target entity, we can rank the
type candidates using the following function:
\begin{align*}
\hat{t} = \argmax_{\hat{t}\in \mathcal{T}} \frac{1}{|\mathcal{N}|} 
\left\{ \sum_{(\Tilde{r}, \Tilde{t})\in \mathcal{N}} p(s_t 
= \hat{t}|r = \Tilde{r}, o_t = \Tilde{t})+\sum_{(\Tilde{r}, \Tilde{t})\in \mathcal{N}} p(o_t 
= \hat{t}|r = \Tilde{r}, s_t = \Tilde{t}) \right\},
\end{align*}
where $\mathcal{N}$ denotes the set of all $(\Tilde{r}, \Tilde{t})$
combinations in the target entity's neighborhood. 

\subsubsection{Classifier-based Method.} We also explore the node
classification approach to solve the type prediction problem as a
benchmarking method.  Specifically, we experiment with the combination
of pretrained TransE embedding as entity features and use SVM and
XGBoost as classifiers. To get the pretrained TransE embedding for
FB15k, we set the batch size to 1000, negative sample size to 256,
hidden dimension to 1000, $\alpha_1 = 1$, the margin parameter $\gamma_1
= 24$, and the learning rate $\eta_1 = 0.0001$, and train for 150000
epochs.  For the SVM classifier, we set the regularization parameter
$C=1$ and adopt the Radial Basis Function (RBF) kernel. The kernel
coefficient for RBF is $\gamma_{svm} = (\#\text{features} \times
\text{feature variance})^{-1}$. For the XGBoost classifier, we set the
learning rate to $0.1$, the number of estimators to $500$, the maximum
depth of each estimator to $5$, the minimum child weight to 1, and use
the softmax objective function for training. 

\subsection{Experimental Results}

To evaluate the performance of the CORE method and several benchmarking
methods, we use the Mean Reciprocal Rank (MRR) and Hits@k as the
performance metrics. Since models are trained to favor observed types as
top choices, we filter the observed types from all possible type
candidates when computing the ranking based on a scoring function. 

First, we show the MRR and Hits@k results for RotatE and ComplEx with
different type dimensions for FB15k-ET in Fig.  \ref{RotatE_dim} and Fig.
\ref{ComplEx_dim}, respectively. The optimal type dimensions for the
RotatE and the ComplEx embeddings are 700 and 550. We adopt
this setting in the following experiments.

Table \ref{fbyago} shows the results for FB15k-ET and YAGO34k-ET. We see
from the table that our proposed CORE models offer the state-of-the-art
performance in both datasets. CORE-ComplEx achieves the best performance
in all categories except Hits@1 for FB15K-ET.  It outperforms the
previous best method, ConnectE-E2T-TRT, by a significant margin for
YAGO34K-ET dataset. 

We also experiment with classifier-based methods and observe that
classifier-based methods do not scale well for large datasets such as FB15k-ET and YAGO43k-ET. The training
time for classifier grows significantly with the size of the label set
and the feature dimension. In addition, the performance of the
classifier-based method is much lower than the statistical-based and
embedding-based methods. 

On the other hand, for datasets with a small number of distinct types,
classifier-based methods do have some advantage. Table \ref{DB111K174}
compares the performance of several type prediction methods for
DB111K-174. The SVM classifier with pretrained TransE embedding features
outperforms all other methods. The idea to incorporate hierarchical structural information of types by JOIE does not seem effective for type prediction since even the simple statistical-based
method such as SDType can outperform the JOIE baseline for MRR and Hits@1 metrics. By conditioning
on neighbor type label information, SDType-Cond can further boost the
performance. The performance gap among different benchmarking methods
is smaller on DB111K-174.

To gain insights into entity type prediction, we provide an illustrative example of type prediction in the YAGO43k-ET dataset. In Table \ref{actual_type_prediction}, we compare the top three type predictions by ConnectE and CORE for some well-known people and place. Although there are some lapses in CORE's predictions, the model can make right decisions for most queries and majority of the top three candidates are in fact valid type labels for the corresponding target entity. These results demonstrate the impressive prediction power of our proposed CORE model, given the enormous amount of unique type labels.

\section{Conclusion and Future Work}\label{sec:conclusion}

A complex regression and embedding method, called CORE, has been proposed to
solve entity type prediction problem in this work by exploiting the
expressive power of RotatE and ComplEx models. It embeds entities and
types in two different complex spaces and used a complex regression
model to measure the relatedness of entities and types. Finally, it optimizes embedding and
regression parameters jointly to reach the optimal performance under
this framework. Experimental results demonstrate that CORE
offers great performance on representative KG entity type
inference datasets and outperforms the state-of-the-art model by a significant margin for YAGO34K-ET dataset.

There are several research directions worth exploration in the future.
First, the use of textual descriptions and transformer models to extract
features for entity type prediction can be investigated. Second, we can
examine the multilabel classification framework for entity type
prediction since it shares a similar problem formulation.  Although both
try to predict multiple target labels, there are however differences.
For multilabel classification, objects in the train set and the test set
are disjoint. That is, we train the classifier using the train set and
test it on a different set. For entity type prediction, the two sets are
not disjoint. In training, a model is trained with a set of entity
feature vectors and their corresponding labels. In inference, it is
often to infer missing type labels for the same set of entities. Third,
a binary classifier can also be used for entity type prediction. Yet,
there exist far more negative samples than positive ones, and it
requires good selection of negative examples to handle the data imbalance
problem.

\bibliographystyle{unsrt}  
\bibliography{references}

\end{document}